\documentclass{article}
\usepackage{spconf,amsmath,graphicx}
\usepackage{amsmath,amsfonts}
\usepackage{algorithmic}
\usepackage{algorithm}
\usepackage{array}
\usepackage{subfigure}
\usepackage{graphicx}
\usepackage{multicol}
\usepackage{multirow}
\usepackage{textcomp}
\usepackage{stfloats}
\usepackage{url}
\usepackage{verbatim}
\usepackage{graphicx}
\usepackage{cite}
\usepackage{booktabs}
\usepackage{tabularx}
\usepackage{amsfonts,amssymb}
\usepackage{multirow}
\usepackage{amsmath,amsfonts}
\usepackage{algorithmic}
\usepackage{algorithm}
\usepackage{array}
\usepackage{bbding}
\usepackage{wrapfig}
\usepackage[table]{xcolor}

\title{RSFDM-Net: Real-time Spatial and Frequency Domains Modulation Network for Underwater Image Enhancement
}

\name{Jingxia Jiang$^{1}$, Jinbin Bai$^{2}$,
 Yun Liu$^{3}$, Junjie Yin$^{1}$, Sixiang Chen$^{1}$, Tian Ye$^{1}$, Erkang Chen$^{1, 4*}$
 \thanks{$^*$Corresponding author: ekchen@jmu.edu.cn. This work was supported in part by Natural Science Foundation of Fujian Province (2021J01867), Xiamen Municipal Bureau of Ocean Development (22CZB013HJ04).}}

\address{$^1$School of Ocean Information Engineering, Jimei University, Xiamen, China \\
$^2$Department of Computer Science, National University of Singapore, Singapore \\
$^3$College of Artificial Intelligence, Southwest University, Chongqing, China \\
$^4$Fujian Provincial Key Laboratory of Oceanic Information Perception and Intelligent Processing, China}
\vspace{-1.5cm}

\begin{document}

\small
%
\maketitle
\begin{abstract}
Underwater images typically experience mixed degradations of brightness and structure caused by the absorption and scattering of light by suspended particles. To address this issue, we propose a Real-time Spatial and Frequency Domains Modulation Network (RSFDM-Net) for the efficient enhancement of colors and details in underwater images. Specifically, our proposed conditional network is designed with Adaptive Fourier Gating Mechanism (AFGM) and Multiscale Convolutional Attention Module (MCAM) to generate vectors carrying low-frequency background information and high-frequency detail features, which effectively promote the network to model global background information and local texture details. To more precisely correct the color cast and low saturation of the image, we introduce a Three-branch Feature Extraction (TFE) block in the primary net that processes images pixel by pixel to integrate the color information extended by the same channel (R, G, or B). This block consists of three small branches, each of which has its own weights. Extensive experiments demonstrate that our network significantly outperforms over state-of-the-art methods in both visual quality and quantitative metrics.

\vspace{0.1cm}
 
.
\end{abstract}

\vspace{-0.6cm}
\section{Introduction}
\label{sec:intro}
Underwater imaging plays a significant role in underwater robotics \cite{wang2020real}, providing essential information for perceiving and understanding underwater environments. Recently, more and more works~\cite{10.1007/978-3-031-19800-7_8,Liu_2022_CVPR,liu2022single,Ye_2022_ACCV,chen2022msp} have paid attention to realw-world image restoration problems with challenging degradations. According to the Jaffe-McGlamey imaging model~\cite{mcglamery1980computer,jaffe1990computer}, underwater imaging consists of a linear superposition of direct, back scattered, and forward scattered components. In general, the effects of forward scattering are negligible, thus the imaging model can be simplified as:
\begin{equation}
   \mathbf{I}_\mathbf{c}(\mathbf{x})=\mathbf{J}_\mathbf{c}(\mathbf{x}) t_\mathbf{c}(\mathbf{x})+\mathbf{B_\mathbf{c}}(1-t_\mathbf{c}(\mathbf{x})), c\in\left\{R,G,B\right\}
\label{eq:underwater imaging model}
\end{equation}
 where $\mathbf{I}_\mathbf{c}(\mathbf{x})$ is the observed intensity in the color channel $c$ of the input image at the pixel $\mathbf{x}$, $\mathbf{J}_\mathbf{c}(\mathbf{x})$
represents the restored image, $\mathbf{B_\mathbf{c}}$ represents the background light, and $t_\mathbf{c}(\mathbf{x})$ is the transmission map, where $c$ represents the red, green, and blue channels.
\begin{equation}
    t_\mathbf{c}(\mathbf{x}) = e^{-\beta_\mathbf{c} d(x)}
\end{equation}
where $d(x)$ is the distance from the camera to the radiant object, and $\beta_\mathbf{c}$ is the spectral volume attenuation coefficient.
Original underwater images are often impacted by long-distance backscattering, selective absorption, and light scattering, resulting in low contrast, low brightness, significant chromatic aberration, blurred details, and uneven bright spots. Therefore, how to improve the visual quality of underwater images is a challenging task.

In the early years, some works\cite{Drews_2013_ICCV_Workshops, peng2017underwater, song2020enhancement, chen2022robust} were developed to address chromatic aberration and blurring effects in underwater images. However, these traditional methods lack the flexibility needed for implementation. For example, an inaccurate estimation of intermediate parameters or invalid underwater optical properties could lead to unsatisfactory results. To overcome this limitation, Li et al.~\cite{li2020underwater} simulate realistic underwater images according to different water types and underwater imaging physical models. Wang et al.~\cite{wang2021uiec} propose a dual color space-based convolutional neural network for underwater image enhancement. Jiang et al.~\cite{jiang2022two} design a novel domain-adaptive framework based on transfer learning to convert aerial image deblurring to real-world underwater image enhancement. Although these methods have achieved varying degrees of success, they have yet to design modules in the network to specifically deal with the color shift, low contrast (mainly at low frequencies), and loss of texture details (mainly at high frequencies), respectively. Huo et al.~\cite{huo2021efficient} use wavelet-enhanced learning units to decompose hierarchical features into high-frequency and low-frequency and subsequently enhanced them by normalization and attention mechanisms. While this method was highly successful, its larger amount of network parameters (6.3M) and computational requirements (0.22s to enhance a 720P underwater image) are not suitable for existing underwater equipment.

According to research \cite{jamadandi2019exemplar}, the task of underwater image enhancement has remarkable similarities with style transfer: both tasks are essentially concerned with changing the style (such as color) of an image while preserving its thematic content. However, compared to style transfer, underwater images pay more attention to the description of texture details, making the processing conditions even more demanding.

\begin{figure*}[t!]
    \centering
    \vspace{-1.2cm}
    \includegraphics[width=0.6\textwidth]{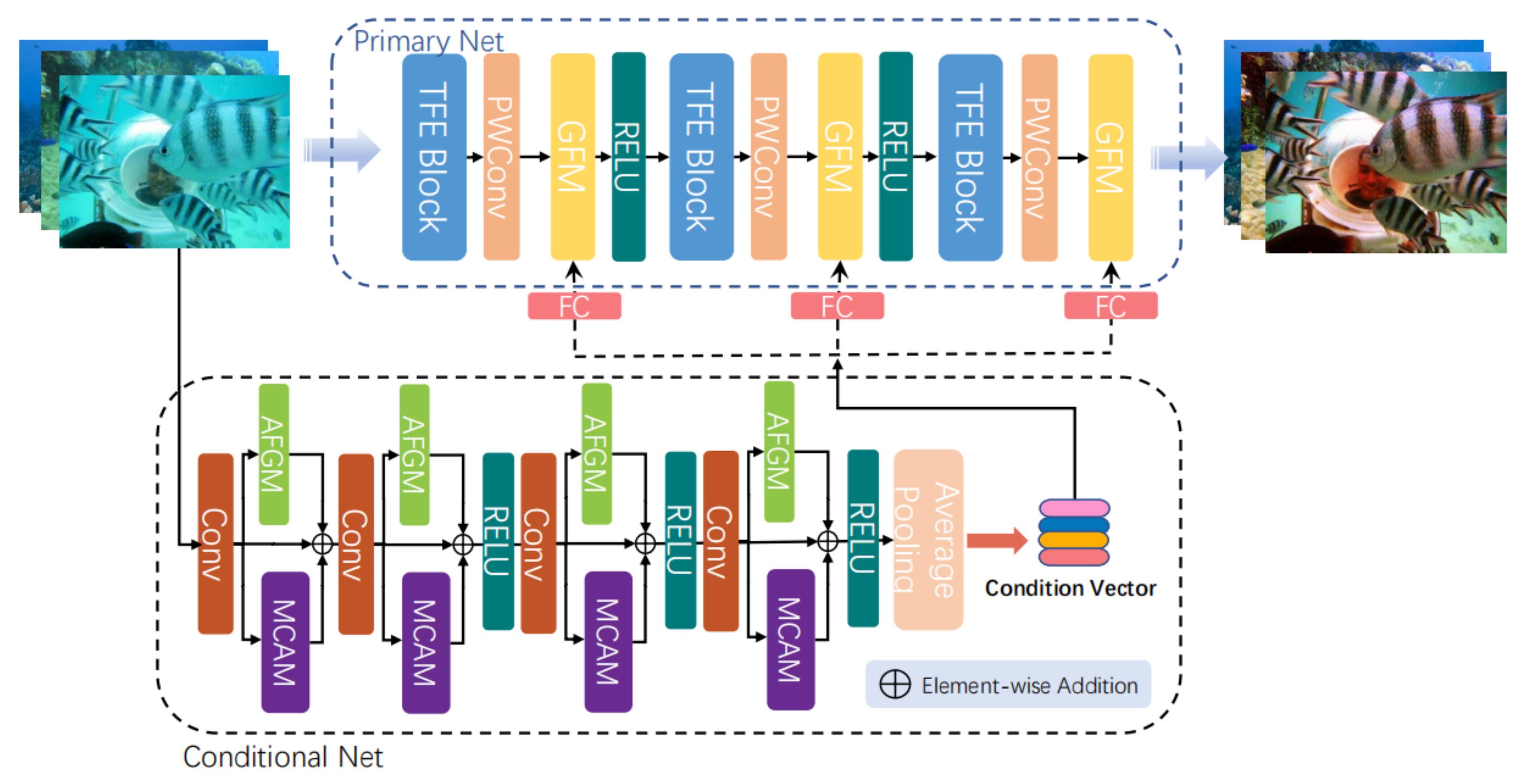}
    \caption{\small Overview of our RSFDM-Net.The input enters both the primary and conditional nets for processing, and then the conditional vectors are interpolated into different layers of the primary net by GSM to guidingly control image enhancement.}
    \label{fig:MAIN}\vspace{-0.3cm}
\end{figure*}

In this work, drawing inspiration from style transfer techniques, we propose the Real-time Spatial and Frequency Domains Modulation Network (RSFDM-Net) for efficient color and detail enhancement in underwater images. It comprises a conditional net with solid guidance and Global Style Modulation (GSM) that can modulate the style of different layers of the primary net. Due to the different light attenuation rates of different wavelengths in an underwater environment, the red light attenuates the fastest when propagating in water, and the blue-green light attenuates the slowest; thus, the difference between the R, G, and B channels of the underwater image is significant \cite{jiang2022underwater}. Previous methods \cite{9120029, han2022underwater} only regard a single channel as the smallest unit of instance normalization, but they ignore the correlation between the channels: all the channels in the network are actually obtained by expanding on the three channels of R, G, and B. In response to this, we design TFE in the primary net that processes images pixel by pixel. The three-branch feature extraction block can more effectively capture the information of color features extended by R, G, or B channel. Many deep learning-based frameworks~\cite{li2020underwater, wang2021uiec, fu2022uncertainty} exploit different kinds of attention modules in the spatial domain to capture high-frequency texture information of images while ignoring low-frequency information which is equally important in hybrid degradation. Inspired by this, we develop AFGM and MCAM based on Conditional Sequence  Retouching Network (CSRNet)~\cite{he2020conditional}. Unlike CSRNet~\cite{he2020conditional} focusing on global retouching, the proposed targeted module can more efficiently and detailedly facilitate the network to model global background information and local texture details. Specifically, the conditional network uses AFGM to capture the global background information contained in the magnitude component and fuses with the multi-scale image details obtained by MCAM, which forces the conditional net to pay attention to both low-frequency contrast problems and loss of high-frequency detail. Finally, the conditional net output after average pooling is used as a conditional vector. The vector carrying low-frequency background information and high-frequency detail features are converted into learnable parameters by GSM, which are broadcast to different layers of the primary net to guidingly control image enhancement. Extensive experiments demonstrate that RSFDM-Net achieves a substantial improvement compared with state-of-the-art methods. RSFDM-Net also lays a better solid foundation for practicality, which can enhance underwater images of size 720P in real-time. Our main contributions in this paper are summarized as follows:
\begin{itemize}
    \item We propose a Real-time Spatial and Frequency Domains Modulation Network (RSFDM-Net) for efficient color and detail enhancement in underwater. Among it, the conditional vector is transformed into learnable parameters via global style modulation, which is broadcasted to different layers of the primary net to guidingly control the progressive enhancement of images.
    
    \item We present a conditional net with AFGM and MCAM. This strongly guided net can generate conditional vectors carrying low-frequency background information and high-frequency detail features. It aims to more effectively promote our network to model global background information and local texture details.

    \item We design TFE block to correct the color cast of underwater images by integrating the color information extended by the same channel (R, G, or B).
\end{itemize}\vspace{-0.25cm}
\vspace{-0.1cm}
\section{Methodology}
\vspace{-0.1cm}
Underwater images commonly experience mixed degradation in the form of color shift, low contrast (particularly at low frequencies), and lack of texture details (particularly at high frequencies). Our method is aimed at quickly and efficiently enhancing images with minimal computational cost. Drawing inspiration from style transfer techniques, we present a Real-time Spatial and Frequency Domains Modulation Network (RSFDM-Net) for effective enhancement of colors and details in underwater images.
\vspace{-0.4cm}
\subsection{Real-time Defined Underwater Image Enhancement Network}
\vspace{-0.2cm}
\begin{figure}[!t]
\centering
\includegraphics[width=0.29\textwidth]{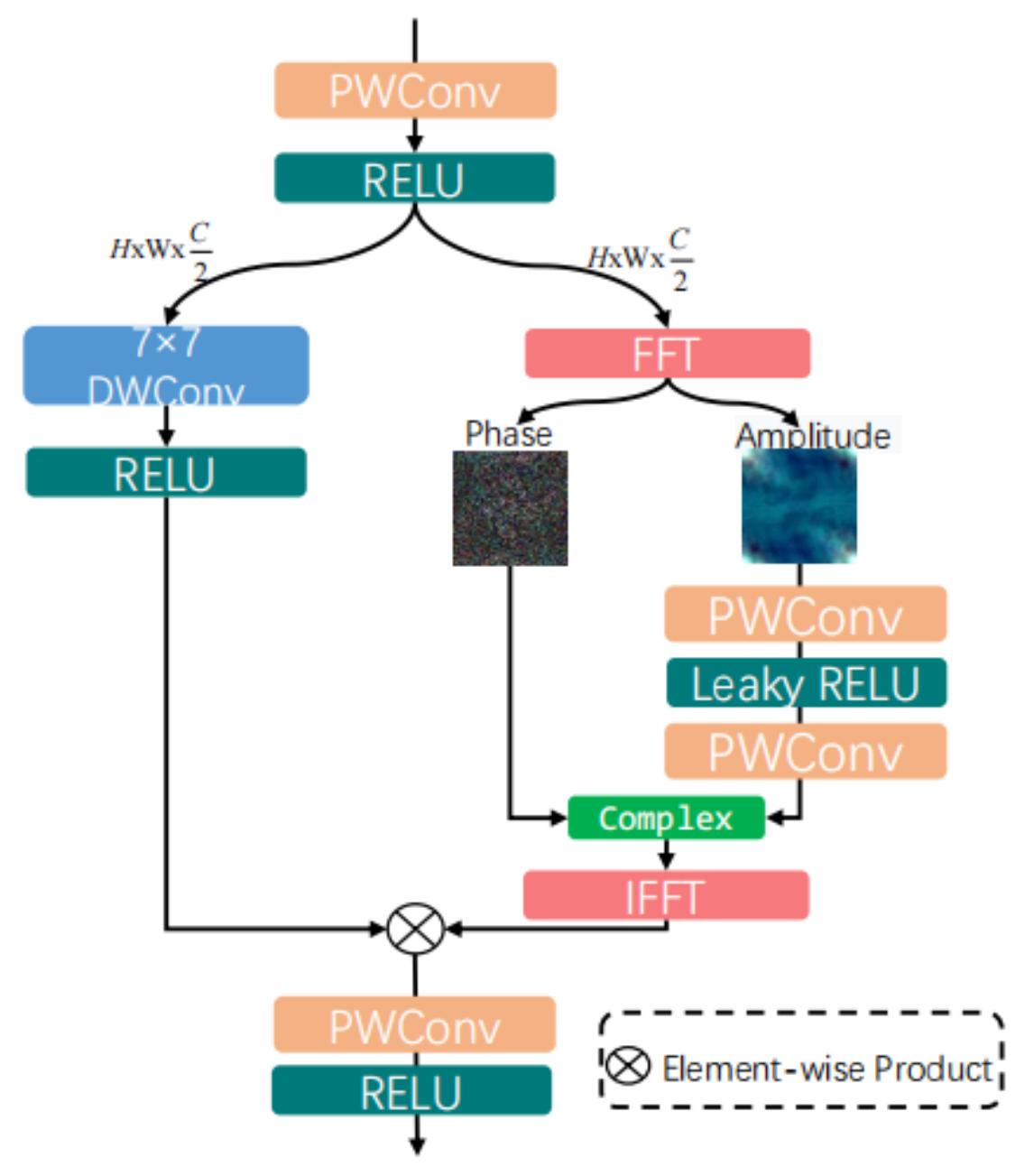} 
\caption{\small 
Details of AFGM.}\label{fig:FIG2}\vspace{-0.4cm}
\end{figure}
The whole pipeline of RSFDM-Net is illustrated in Fig.\ref{fig:MAIN}. The proposed network structure contains a primary net and a conditional net. The primary net takes underwater degraded images as input and generates enhanced images, where TFE blocks help to correct the underwater images' color cast. To further guide the primary net to better enhance degraded images, a conditional net is proposed to cooperate with the primary net. It can generate conditional vectors which carry low-frequency background information and high-frequency detail features through AFGM and MCAM, helping to more efficiently model the global background information and local texture details. GSM then converts the conditional vectors into learnable parameters, which are broadcasted to different layers of the primary net to guide underwater image enhancement. GSM adopts scaling and shifting operations to modulate intermediate style of the underwater image. The operation of GSM is described as follows:
\begin{equation}
GSM({x_i}) = \gamma *{x_i} + \beta ,0 < i \le N
\end{equation}
 where $\gamma$, $\beta$ are affine parameters.
\vspace{-0.3cm}
\subsection{Adaptive Fourier Gating Mechanism}

Methods\cite{li2020underwater, wang2021uiec, fu2022uncertainty} restore the degraded image in the spatial domain alone. According to \cite{katznelson2004introduction}, processing information in Fourier
space is capable of capturing the global frequency representation in the frequency domain. In contrast, the normal convolution focuses on learning local representations in the spatial domain. In this way, we propose AFGM illustrated in Fig.\ref{fig:FIG2} to learn more representative features. Specifically, the module separates the magnitude component carrying global information through a series of frequency domain operations, which include the Fourier transform.
\vspace{-0.2cm}
\subsection{Multi-scale Convolutional Attention Module}
\begin{figure}[!t]
\centering
\includegraphics[width=0.29\textwidth]{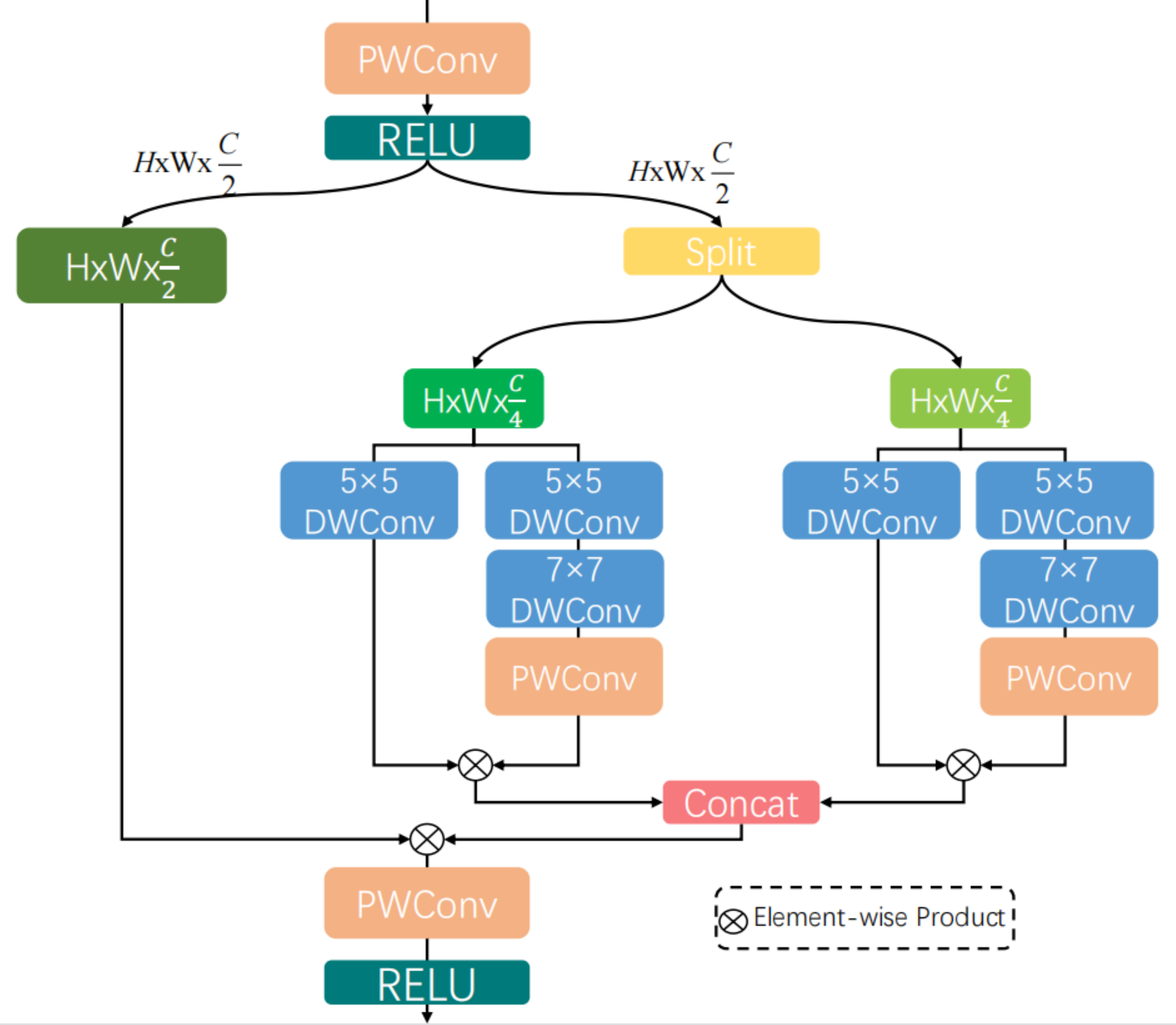} 
\caption{\small 
Two sets of multi-scale convolution are included in MCAM to obtain a promising extraction capability.}\label{fig:FIG3}\vspace{-0.2cm}
\end{figure}

We hope the vector can store global information and texture details, so as to generate a conditional vector with abundant scene knowledge. As shown in Fig.\ref{fig:FIG3}, MCAM is adopted in the conditional net for learning attention maps with degraded local details. Specifically, given the feature $X_{ci}$ after split operation, $X_{ci} \in {R^{C{\rm{ \times }}H{\rm{ \times }}W}}$. The key process of MCAM is formulated as:
\begin{equation}
\begin{array}{l}
{X_{g1i}} = {f_{PW}}({f_{DWD}}({f_{DW}}({X_{ci}}))),i=1,2\\
{X_{g2i}} = {f_{DW}}({X_{ci}}),i=1,2\\
\end{array}
\end{equation}
where $f_{DW}( \cdot )$ is a $(2d-1)$ $\times$ $(2d-1)$ depth-wise convolution and $f_{DWD}(\cdot )$ is a [$\frac{K}{d} \times \frac{K}{d}$] depth-wise d-dilation convolution. ${f_{PW}} (\cdot)$  represents a point-wise convolution.
\vspace{-0.2cm}
\subsection{Three-branch Feature Extraction}

\begin{figure}[!t]
\centering
\includegraphics[width=0.28\textwidth]{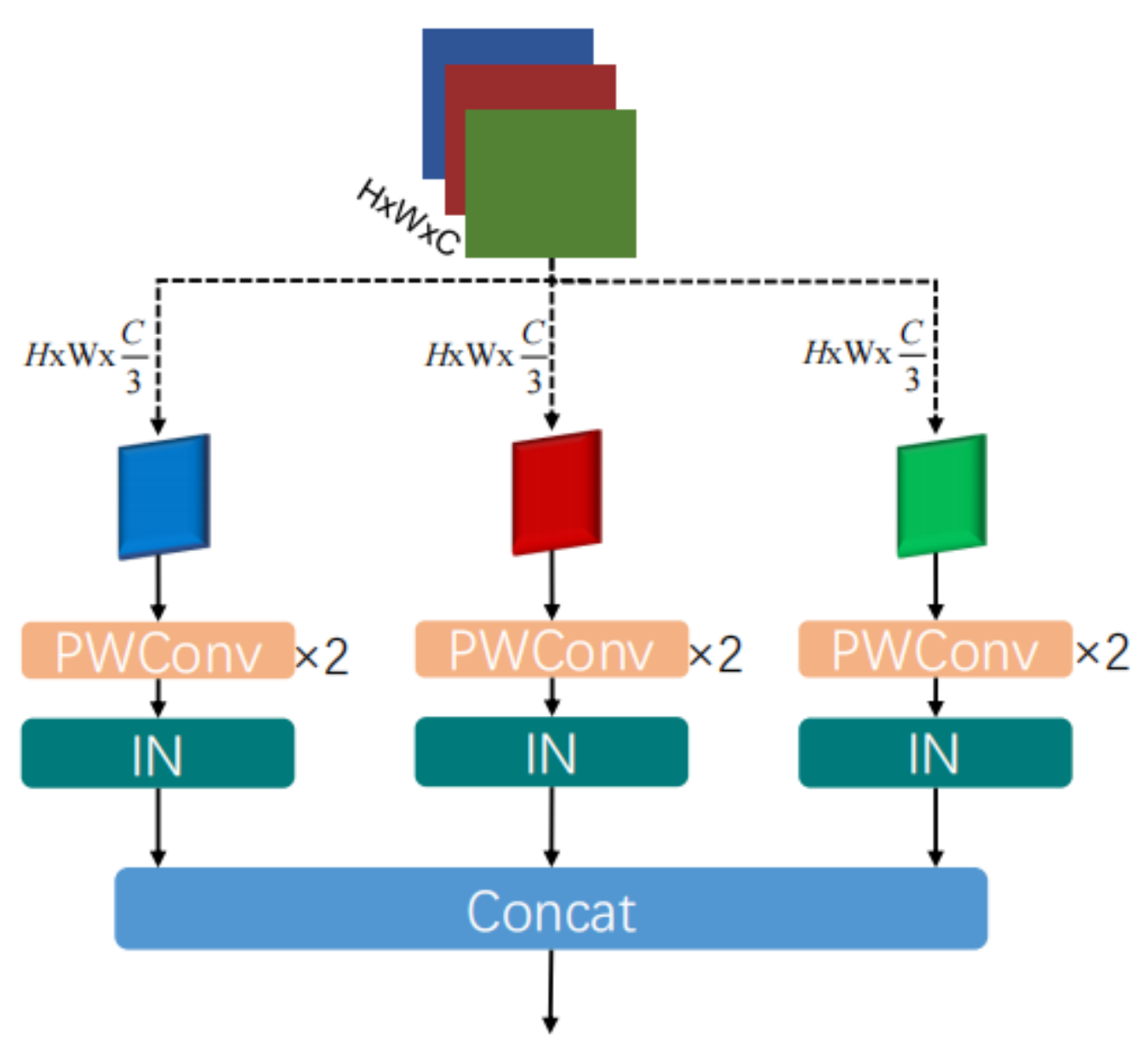}
\caption{\small 
First, integrate and expand the color features through the convolution operation, then incorporate the color information extended from the same channel (R or G or B) through channel grouping normalization, and finally, stitch the output results of the three branches together.}\label{fig:FIG4}\vspace{-0.4cm}
\end{figure}

Previous methods \cite{9120029,han2022underwater} ignore the correlation between channels: all channels in the network are derived from the three channels of R, G, and B. To address this limitation, we design the TFE block to better capture the distribution of color features on R, G, and B channels. As shown in Fig.\ref{fig:FIG4}, the block has three small branches, and the weights are not shared between each branch. The effectiveness of this block is demonstrated in ablation study.
\vspace{-0.3cm}
\subsection{Loss Function}
We introduce the Charbonnier Loss~\cite{charbonnier1994two} as our basic reconstruction loss:
\begin{equation}
    \mathcal{L}_{rec} = \mathcal{L}_{c}(
    \mathcal{I}(X) ,
    \mathcal{J}_{gt}
    )
\end{equation}
where the $\mathcal{I}$ is our RSFDM-Net, $X$ and $\mathcal{J}_{gt}$ stand for input and ground-truth. $\mathcal{L}_c$ denotes the Charbonnier loss, which can be express as:
\begin{equation}
    \mathcal{L}_{c} =\frac{1}{N} \sum_{i=1}^{N} \sqrt{\left\|X^{i}-Y^{i}\right\|^{2}+\epsilon^{2}}
\end{equation}
where constant $\epsilon$ emiprically set to 1e-3 for all experiments.
In addition, perceptual level of the restored image is also
critical. We apply the perceptual loss to improve the restoration performance. The perceptual loss can be formulated as follows:
\begin{equation}
{\mathcal{L}_{perceptual}} = \sum\limits_{j = 1}^2 {\frac{1}{{{C_j}{H_j}{W_j}}}} ||{\phi _j}(\mathcal{I}(x)) - {\phi _j}(\mathcal{Y})|{|_1}
\end{equation}
wherein the $\phi _j$ represents the specified layer of VGG19 \cite{simonyan2014very}. $C_j$ , $H_j$ , $W_j$ represent the channel number, height, and width of the feature map, relatively.
Overall loss function can be expressed as:
\begin{equation}
\mathcal{L} = {\lambda _1}{\mathcal{L}_c} + {\lambda _2}{\mathcal{L}_{perceptual}}
\end{equation}
where the $\lambda _1$ and $\lambda _2$ are set to 1 and 0.2, relatively.
\vspace{-0.1cm}
\section{Experiments}
\begin{figure}[!t]
\centering
\vspace{-0.4cm}
\includegraphics[width=0.5\textwidth]{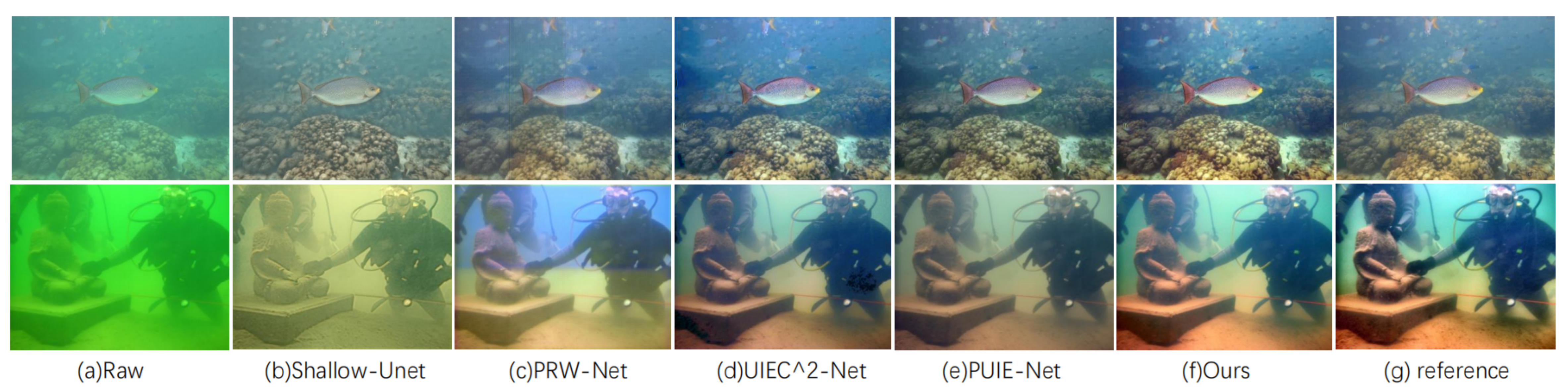} 
\vspace{-0.6cm}
\caption{\small 
Visual comparison with the previous methods on the T90.}\label{fig:ccc}
\vspace{-0.5cm}
\end{figure}
\vspace{-0.15cm}
\subsection{Underwater image enhancement benchmark datasets}
UIEB \cite{8917818} contains 890 high-resolution raw underwater images, corresponding high-quality reference images, and 60 challenge images(C60) for which no corresponding reference images were obtained. The content of the images covers a wide range, such as marine life and divers, where the quality of underwater images significantly degenerate.

\noindent U45 \cite{li2019fusion} includes three degradation types of biased color, low contrast, and haze-like appearance, which has no corresponding reference images.
\vspace{-0.35cm}
\begin{table}[!h]
\centering
\caption{\small{The size of image for testing is 1080$\times$720. The results show that our network can enhance 720P images in real time. In addition, for the trade-off of the amount of parameters and computation, RSFDM-Net also lays a better solid foundation for practicality.}}
\vspace{0.1cm}
\resizebox{8cm}{!}{
\renewcommand\arraystretch{1.1}
\begin{tabular}{c ccccc}

\toprule[1.2pt]
\rowcolor[HTML]{EFEFEF} 
\textbf{Method}   & PRWNet\cite{9607639}  & Shallow-UWnet\cite{naik2021shallow} & UIEC\textasciicircum{}2-Net\cite{wang2021uiec} & PUIE-Net\cite{fu2022uncertainty} & Ours  \\\hline
\textbf{\#Param}  & 6.30M   & \underline{219.46K}       & 534.96K  & 1.4M     & \textbf{110.95K} \\
\textbf{\#GFLOPs} & \underline{223.37G} & 304.75G       & 367.53G  & 423.05G  & \textbf{31.33G}  \\ 
\textbf{\#Runtime} & 0.216s & \textbf{0.032s}       & 0.174s  & 0.070s  & \underline{0.039s}  \\ \hline
\bottomrule[1.2pt]\label{table:222}
\end{tabular}}
\vspace{-0.5cm}
\end{table}
\vspace{-0.4cm}
\subsection{Experimental implementation details}
\noindent\textbf{Training Details.} We implemented our network using PyTorch with a single RTX 3090 GPU. We trained our model for 400 epochs with a patch size of $512\times512$. The Adam optimizer was used with an initial learning rate of $1\times 10^{-4}$. We employed 800 pairs of raw and sharp images selected from the UIEB \cite{8917818} for training, and 90 images, called T90, for testing. We used CyclicLR to adjust the learning rate, with an initial momentum of 0.9 and 0.999. Data augmentation included horizontal flipping, random cropping and randomly rotating the image to $0, 90, 180$ and $270$ degrees.

\noindent\textbf{Testing Detail.} We use the T90 dataset to evaluate the performance of our model in terms of generalization. Additionally, we employ the C60 dataset, which consists of 60 challenge images with no corresponding reference images, as a part of our testing dataset. For further benchmarking, we utilize the U45 \cite{li2019fusion} dataset which contains three degradation types for testing our model.

\noindent\textbf{Evaluation Metrics.} For quantitative measurement, we use the peak signal-to-noise ratio (PSNR) \cite{korhonen2012peak}, structural similarity index metric (SSIM) \cite{wang2004image}, underwater color image quality evaluation (UCIQE) \cite{yang2015underwater}, underwater image quality measure (UIQM) \cite{panetta2015human} and the Natural Image Quality Evaluator (NIQE) \cite{mittal2012making} as objective reference standards for image quality. The PSNR is a full-reference image quality evaluation metric, and it is based on the error between corresponding pixels.  SSIM measures the visual impact of three features of an image: brightness, contrast, and structure. 

\noindent\textbf{Compared with SOTA Methods.} We compare RSFDM-Net with state-of-the-art methods, including nondeep learning and deep learning methods. Nondeep learning methods include UDCP \cite{Drews_2013_ICCV_Workshops}, IBLA \cite{peng2017underwater}, SMBL \cite{song2020enhancement} and MLLE \cite{zhang2022underwater}, and deep learning methods include UWCNN \cite{li2020underwater}, Water-Net \cite{li2019underwater}, PRW-Net \cite{9607639}, Shallow-UWnet \cite{naik2021shallow}, Ucolor \cite{li2021underwater}, UIEC\textasciicircum{}2-Net \cite{wang2021uiec} and the latest PUIE-Net \cite{fu2022uncertainty} method [38]. We can observe that the proposed RSFDM-Net achieves the best results on PSNR and SSIM metrics in Table.\ref{table:111}, which prove our proposed method is good at handling details textures and restore contrast. Compared to the second best approach UIEC\textasciicircum{}2-Net \cite{wang2021uiec}, we exceed the 0.36dB and 0.05 on PSNR and SSIM. Our method also outperforms most methods in no-reference image quality assessment. We present parameters and performance indicators with previous SOTA methods in Table.\ref{table:222}. We also present the visual comparison with previous SOTA methods in Fig.\ref{fig:ccc}. It can be seen that our method can enhance the whole degraded images thoroughly, while the previous methods still have some local plaques.
\vspace{-0.1cm}

\begin{table}[!t]
\vspace{-0.6cm}
\centering
\caption{\small Quantitative comparison of various approaches on the UIEB \cite{8917818} datasets and U45 datasets \cite{li2019fusion}, using PSNR, SSIM, UCIQE, UIQM, and NIQE to evaluate the performance of all methods. Bold and underline indicate the best and second best metrics.↑ represents the higher is the better as well as ↓ represents the lower is the better.}\label{enhancingresults}
\vspace{0.1cm}
\resizebox{8.5cm}{!}{
\renewcommand\arraystretch{1.1}
\begin{tabular}{llllll|lll|lll}
\toprule[1.2pt]
\rowcolor[HTML]{EFEFEF} 
& \multicolumn{5}{c|}{\cellcolor[HTML]{EFEFEF}\textbf{T90\cite{8917818}}}                           & \multicolumn{3}{c|}{\cellcolor[HTML]{EFEFEF}\textbf{C60\cite{8917818}}} & \multicolumn{3}{c}{\cellcolor[HTML]{EFEFEF}\textbf{U45\cite{li2019fusion}}} \\ \cline{2-12} 
\rowcolor[HTML]{EFEFEF} 
\multirow{-2}*{\cellcolor[HTML]{EFEFEF}\textbf{Method}} & PSNR↑           & SSIM↑          & MSE↓           & UCIQE↑         & UIQM↑          & UCIQE↑               & UIQM↑                & NIQE↓       & UCIQE↑             & UIQM↑            & NIQE↓            \\ \hline
(ICCVW'13)UDCP~\cite{Drews_2013_ICCV_Workshops}                                                                 & 13.415          & 0.749          & 0.228          & 0.572          & 2.755          & 0.560                & 1.859                & 5.897       & 0.574              & 2.275            & \underline{4.144}            \\
(TIP'17)IBLA~\cite{peng2017underwater}                                                                   & 18.054          & 0.808          & 0.142          & 0.582          & 2.557          & {\textbf {0.584}}          & 1.662                & 5.954       & 0.565              & 2.387            & \textbf{4.100}            \\
(TIP'19)WaterNet~\cite{li2019underwater}                                                               & 16.305          & 0.797          & 0.161          & 0.564          & 2.916          & 0.550                & 2.113                & 5.778       & 0.576              & 2.957            & 5.346            \\
(TB'20)SMBL~\cite{song2020enhancement}                                                                    & 16.681          & 0.801          & 0.158          & 0.589          & 2.598          & 0.571                & 1.643                & 5.891       & 0.571              & 2.387            & 4.412            \\
(PR'20)UWCNN~\cite{li2020underwater}                                                                   & 17.949          & 0.847          & 0.221          & 0.517          & 3.011          & 0.492                & 2.222      & 5.837       & 0.527              & 3.063            & 4.187            \\
(ICCVW'20)PRW-Net~\cite{9607639}                                                                   & 20.787          & 0.823          & 0.099          & \underline{0.603}          & \textbf{3.062}         & 0.572             & \textbf{2.717}       & \underline{5.425}       & \underline{0.625}              & 3.026            & 4.157            \\
(AAAI'21)Shallow-UWnet~\cite{naik2021shallow}                                                         & 18.278          & 0.855          & 0.131          & 0.544          & 2.942          & 0.521                & 2.212                &     5.874        &         0.545           &       3.109           &   4.241               \\
(TIP'21)Ucolor~\cite{li2021underwater}                                                                 & 21.093          & 0.872          & 0.096          & 0.555          & {\underline{3.049}}    & 0.530                & 2.167                & 6.298       & 0.554              & \underline{3.148}            & 4.687            \\
(SPIC'21)UIEC\textasciicircum{}2-Net~\cite{wang2021uiec}                                           & \underline{22.958} & \underline{0.907} & \underline{0.078} & 0.599 & 2.999          & 0.580     & {\underline{2.280}}          & 5.438       & 0.604              & 3.125            & 4.182            \\
(TIP'22)MLLE~\cite{zhang2022underwater}                                                                   & 19.561          & 0.845          & 0.115          & 0.592          & 2.624          & 0.581                & 1.977                &\textbf{4.925}       & 0.597              & 2.454            & 4.725            \\
(ECCV'22)PUIE-Net(MC)~\cite{fu2022uncertainty}                                                              & 21.382          & 0.882          & 0.093          & 0.566          & 3.021 & 0.543                & 2.155                & 5.935       & 0.563              & \textbf{3.192}            & 4.146            \\ \hline
Ours                                                                           & {\textbf {23.325}}    & {\textbf {0.912}}    & {\textbf{ 0.076}}    & {\textbf {0.616}}    & 2.678          & \underline{0.582}       & 1.940                &    5.842         &     {\textbf { 0.627}}   &    2.995              &     4.179         \\ 
\bottomrule[1.2pt]\label{table:111}
\vspace{-1cm}
\end{tabular}}
\end{table}
\vspace{-0.2cm}
\subsection{Ablation Study}
For ablation studies, we follow the basic settings presented above and conduct experiments to demonstrate the effectiveness of the components of our proposed comprehensive manner. The results are listed on the Table.\ref{table:333}. In the \textbf{S1} setting, the Model consists of original CSR-Net\cite{he2020conditional}. Based on top of \textbf{S1}, we applied the other components to form different settings: \textbf{ii} (Apply TFE) and \textbf{iii} (Apply the Instance Normalization). It can be seen that the inclusion of TFE blocks improve the enhancing performance of the network better than IN. Secondly, the effectiveness of AFGM and MCAM are also verified by ablation study, as listed on the bottom rows of Table.\ref{table:333}. Based on setting \textbf{ii}, we extra apply AFGM in network as setting \textbf{iv}. Similarly, we have added MCAM as setting \textbf{v} and have applied CA as setting \textbf{vi}. We find AFGM greatly improves the enhancement performance in terms of PSNR. Additionally, MCAM proves more beneficial in restoring image details than when it is replaced with CA.
\vspace{-0.45cm}
\begin{table}[!h]
\centering
\caption{\small{Ablation studies on main modules of RSFDM-Net. IN and CA stand for Instance Normalization and Channel Attention module.}}
\vspace{0.2cm} 
\resizebox{5cm}{!}{
\renewcommand\arraystretch{1.1}
\begin{tabular}{
ccll}
\hline
\rowcolor[HTML]{EFEFEF} 
\multicolumn{1}{l}{\cellcolor[HTML]{EFEFEF}\textbf{Setting}} & \textbf{Model} & \textbf{PSNR↑} & \textbf{SSIM↑} \\ \hline
$S1$                                                     & $S1$           & 22.374         & 0.874          \\
ii                                                           & $S1$+GCN       & 22.729         & 0.891          \\
iii                                                          & $S1$+IN        & 22.677         & 0.887          \\
iv                                                           & $S1$+GCN+AFGM  & 23.101         & 0.896          \\
v                                                            & $S1$+GCN+MCAM  & 23.065         & 0.903          \\
vi                                                           & $S1$+GCN+CA    & 22.959         & 0.886          \\
vii                                                          & RSFDM-Net(Ours) & \textbf{23.325}         & \textbf{0.912 }         \\ \hline
\bottomrule[1.2pt]\label{table:333}
\vspace{-0.5cm}\end{tabular}}
\vspace{-0.5cm}
\end{table}
\vspace{-0.2cm}
\section{CONCLUSIONS}
\vspace{-0.2cm}
We propose a Real-time Spatial and Frequency Domains Modulation Network (RSFDM-Net) for efficient color and detail enhancement in underwater images. Moreover, We present a conditional net with Adaptive Fourier Gate Mechanism (AFGM) and Multiscale Convolutional Attention Module (MCAM) to more effectively promote the network to model global background information and local texture details. To detailedly correct the color cast and low saturation of the image, we design Three-branch Feature Extraction (TFE) block to achieve it by integrating the color information extended by the same channel. Extensive experiments on underwater datasets demonstrate the superiority of our RSFDM-Net. 


\vfill\pagebreak

\bibliographystyle{IEEEbib}
\footnotesize{\bibliography{strings,refs}}

\begin{thebibliography}{10}

\bibitem{wang2020real}
Yu~Wang, Chong Tang, Mingxue Cai, Jiye Yin, Shuo Wang, Long Cheng, Rui Wang,
  and Min Tan,
\newblock ``Real-time underwater onboard vision sensing system for robotic
  gripping,''
\newblock {\em IEEE Transactions on Instrumentation and Measurement}, vol. 70,
  pp. 1--11, 2020.

\bibitem{10.1007/978-3-031-19800-7_8}
Tian Ye, Yunchen Zhang, Mingchao Jiang, Liang Chen, Yun Liu, Sixiang Chen, and
  Erkang Chen,
\newblock ``Perceiving and modeling density for image dehazing,''
\newblock in {\em Computer Vision -- ECCV 2022}, Shai Avidan, Gabriel Brostow,
  Moustapha Ciss{\'e}, Giovanni~Maria Farinella, and Tal Hassner, Eds., Cham,
  2022, pp. 130--145, Springer Nature Switzerland.

\bibitem{Liu_2022_CVPR}
Yun Liu, Zhongsheng Yan, Aimin Wu, Tian Ye, and Yuche Li,
\newblock ``Nighttime image dehazing based on variational decomposition
  model,''
\newblock in {\em Proceedings of the IEEE/CVF Conference on Computer Vision and
  Pattern Recognition (CVPR) Workshops}, June 2022, pp. 640--649.

\bibitem{liu2022single}
Yun Liu, Zhongsheng Yan, Tian Ye, Aimin Wu, and Yuche Li,
\newblock ``Single nighttime image dehazing based on unified variational
  decomposition model and multi-scale contrast enhancement,''
\newblock {\em Engineering Applications of Artificial Intelligence}, vol. 116,
  pp. 105373, 2022.

\bibitem{Ye_2022_ACCV}
Tian Ye, Sixiang Chen, Yun Liu, Yi~Ye, Jinbin Bai, and Erkang Chen,
\newblock ``Towards real-time high-definition image snow removal: Efficient
  pyramid network with asymmetrical encoder-decoder architecture,''
\newblock in {\em Proceedings of the Asian Conference on Computer Vision
  (ACCV)}, December 2022, pp. 366--381.

\bibitem{chen2022msp}
Sixiang Chen, Tian Ye, Yun Liu, Taodong Liao, Yi~Ye, and Erkang Chen,
\newblock ``Msp-former: Multi-scale projection transformer for single image
  desnowing,''
\newblock {\em arXiv preprint arXiv:2207.05621}, 2022.

\bibitem{mcglamery1980computer}
BL~McGlamery,
\newblock ``A computer model for underwater camera systems,''
\newblock in {\em Ocean Optics VI}. SPIE, 1980, vol. 208, pp. 221--231.

\bibitem{jaffe1990computer}
Jules~S Jaffe,
\newblock ``Computer modeling and the design of optimal underwater imaging
  systems,''
\newblock {\em IEEE Journal of Oceanic Engineering}, vol. 15, no. 2, pp.
  101--111, 1990.

\bibitem{Drews_2013_ICCV_Workshops}
P.~Drews, Jr., E.~do~Nascimento, F.~Moraes, S.~Botelho, and M.~Campos,
\newblock ``Transmission estimation in underwater single images,''
\newblock in {\em Proceedings of the IEEE International Conference on Computer
  Vision (ICCV) Workshops}, June 2013.

\bibitem{peng2017underwater}
Yan-Tsung Peng and Pamela~C Cosman,
\newblock ``Underwater image restoration based on image blurriness and light
  absorption,''
\newblock {\em IEEE transactions on image processing}, vol. 26, no. 4, pp.
  1579--1594, 2017.

\bibitem{song2020enhancement}
Wei Song, Yan Wang, Dongmei Huang, Antonio Liotta, and Cristian Perra,
\newblock ``Enhancement of underwater images with statistical model of
  background light and optimization of transmission map,''
\newblock {\em IEEE Transactions on Broadcasting}, vol. 66, no. 1, pp.
  153--169, 2020.

\bibitem{chen2022robust}
Sixiang Chen, Erkang Chen, Tian Ye, and Chenghao Xue,
\newblock ``Robust back-scattered light estimation for underwater image
  enhancement with polarization,''
\newblock {\em Displays}, vol. 75, pp. 102296, 2022.

\bibitem{li2020underwater}
Chongyi Li, Saeed Anwar, and Fatih Porikli,
\newblock ``Underwater scene prior inspired deep underwater image and video
  enhancement,''
\newblock {\em Pattern Recognition}, vol. 98, pp. 107038, 2020.

\bibitem{wang2021uiec}
Yudong Wang, Jichang Guo, Huan Gao, and Huihui Yue,
\newblock ``Uiec\^{} 2-net: Cnn-based underwater image enhancement using two
  color space,''
\newblock {\em Signal Processing: Image Communication}, vol. 96, pp. 116250,
  2021.

\bibitem{jiang2022two}
Qun Jiang, Yunfeng Zhang, Fangxun Bao, Xiuyang Zhao, Caiming Zhang, and Peide
  Liu,
\newblock ``Two-step domain adaptation for underwater image enhancement,''
\newblock {\em Pattern Recognition}, vol. 122, pp. 108324, 2022.

\bibitem{huo2021efficient}
Fushuo Huo, Bingheng Li, and Xuegui Zhu,
\newblock ``Efficient wavelet boost learning-based multi-stage progressive
  refinement network for underwater image enhancement,''
\newblock in {\em Proceedings of the IEEE/CVF International Conference on
  Computer Vision}, 2021, pp. 1944--1952.

\bibitem{jamadandi2019exemplar}
Adarsh Jamadandi and Uma Mudenagudi,
\newblock ``Exemplar-based underwater image enhancement augmented by wavelet
  corrected transforms,''
\newblock in {\em Proceedings of the IEEE/CVF Conference on Computer Vision and
  Pattern Recognition Workshops}, 2019, pp. 11--17.

\bibitem{jiang2022underwater}
Qiuping Jiang, Yuese Gu, Chongyi Li, Runmin Cong, and Feng Shao,
\newblock ``Underwater image enhancement quality evaluation: Benchmark dataset
  and objective metric,''
\newblock {\em IEEE Transactions on Circuits and Systems for Video Technology},
  vol. 32, no. 9, pp. 5959--5974, 2022.

\bibitem{9120029}
Jing Wang, Ping Li, Jianhua Deng, Yongzhao Du, Jiafu Zhuang, Peidong Liang, and
  Peizhong Liu,
\newblock ``Ca-gan: Class-condition attention gan for underwater image
  enhancement,''
\newblock {\em IEEE Access}, vol. 8, pp. 130719--130728, 2020.

\bibitem{han2022underwater}
Junlin Han, Mehrdad Shoeiby, Tim Malthus, Elizabeth Botha, Janet Anstee, Saeed
  Anwar, Ran Wei, Mohammad~Ali Armin, Hongdong Li, and Lars Petersson,
\newblock ``Underwater image restoration via contrastive learning and a
  real-world dataset,''
\newblock {\em Remote Sensing}, vol. 14, no. 17, pp. 4297, 2022.

\bibitem{fu2022uncertainty}
Zhenqi Fu, Wu~Wang, Yue Huang, Xinghao Ding, and Kai-Kuang Ma,
\newblock ``Uncertainty inspired underwater image enhancement,''
\newblock in {\em Computer Vision--ECCV 2022: 17th European Conference, Tel
  Aviv, Israel, October 23--27, 2022, Proceedings, Part XVIII}. Springer, 2022,
  pp. 465--482.

\bibitem{he2020conditional}
Jingwen He, Yihao Liu, Yu~Qiao, and Chao Dong,
\newblock ``Conditional sequential modulation for efficient global image
  retouching,''
\newblock in {\em Computer Vision--ECCV 2020: 16th European Conference,
  Glasgow, UK, August 23--28, 2020, Proceedings, Part XIII 16}. Springer, 2020,
  pp. 679--695.

\bibitem{katznelson2004introduction}
Yitzhak Katznelson,
\newblock {\em An introduction to harmonic analysis},
\newblock Cambridge University Press, 2004.

\bibitem{charbonnier1994two}
Pierre Charbonnier, Laure Blanc-Feraud, Gilles Aubert, and Michel Barlaud,
\newblock ``Two deterministic half-quadratic regularization algorithms for
  computed imaging,''
\newblock in {\em Proceedings of 1st international conference on image
  processing}. IEEE, 1994, vol.~2, pp. 168--172.

\bibitem{simonyan2014very}
Karen Simonyan and Andrew Zisserman,
\newblock ``Very deep convolutional networks for large-scale image
  recognition,''
\newblock {\em arXiv preprint arXiv:1409.1556}, 2014.

\bibitem{8917818}
Chongyi Li, Chunle Guo, Wenqi Ren, Runmin Cong, Junhui Hou, Sam Kwong, and
  Dacheng Tao,
\newblock ``An underwater image enhancement benchmark dataset and beyond,''
\newblock {\em IEEE Transactions on Image Processing}, vol. 29, pp. 4376--4389,
  2020.

\bibitem{li2019fusion}
Hanyu Li, Jingjing Li, and Wei Wang,
\newblock ``A fusion adversarial underwater image enhancement network with a
  public test dataset,''
\newblock {\em arXiv preprint arXiv:1906.06819}, 2019.

\bibitem{9607639}
Fushuo Huo, Bingheng Li, and Xuegui Zhu,
\newblock ``Efficient wavelet boost learning-based multi-stage progressive
  refinement network for underwater image enhancement,''
\newblock in {\em 2021 IEEE/CVF International Conference on Computer Vision
  Workshops (ICCVW)}, 2021, pp. 1944--1952.

\bibitem{naik2021shallow}
Ankita Naik, Apurva Swarnakar, and Kartik Mittal,
\newblock ``Shallow-uwnet: Compressed model for underwater image enhancement
  (student abstract),''
\newblock in {\em Proceedings of the AAAI Conference on Artificial
  Intelligence}, 2021, vol.~35, pp. 15853--15854.

\bibitem{korhonen2012peak}
Jari Korhonen and Junyong You,
\newblock ``Peak signal-to-noise ratio revisited: Is simple beautiful?,''
\newblock in {\em 2012 Fourth International Workshop on Quality of Multimedia
  Experience}. IEEE, 2012, pp. 37--38.

\bibitem{wang2004image}
Zhou Wang, Alan~C Bovik, Hamid~R Sheikh, and Eero~P Simoncelli,
\newblock ``Image quality assessment: from error visibility to structural
  similarity,''
\newblock {\em IEEE transactions on image processing}, vol. 13, no. 4, pp.
  600--612, 2004.

\bibitem{yang2015underwater}
Miao Yang and Arcot Sowmya,
\newblock ``An underwater color image quality evaluation metric,''
\newblock {\em IEEE Transactions on Image Processing}, vol. 24, no. 12, pp.
  6062--6071, 2015.

\bibitem{panetta2015human}
Karen Panetta, Chen Gao, and Sos Agaian,
\newblock ``Human-visual-system-inspired underwater image quality measures,''
\newblock {\em IEEE Journal of Oceanic Engineering}, vol. 41, no. 3, pp.
  541--551, 2015.

\bibitem{mittal2012making}
Anish Mittal, Rajiv Soundararajan, and Alan~C Bovik,
\newblock ``Making a “completely blind” image quality analyzer,''
\newblock {\em IEEE Signal processing letters}, vol. 20, no. 3, pp. 209--212,
  2012.

\bibitem{zhang2022underwater}
Weidong Zhang, Peixian Zhuang, Hai-Han Sun, Guohou Li, Sam Kwong, and Chongyi
  Li,
\newblock ``Underwater image enhancement via minimal color loss and locally
  adaptive contrast enhancement,''
\newblock {\em IEEE Transactions on Image Processing}, vol. 31, pp. 3997--4010,
  2022.

\bibitem{li2019underwater}
Chongyi Li, Chunle Guo, Wenqi Ren, Runmin Cong, Junhui Hou, Sam Kwong, and
  Dacheng Tao,
\newblock ``An underwater image enhancement benchmark dataset and beyond,''
\newblock {\em IEEE Transactions on Image Processing}, vol. 29, pp. 4376--4389,
  2019.

\bibitem{li2021underwater}
Chongyi Li, Saeed Anwar, Junhui Hou, Runmin Cong, Chunle Guo, and Wenqi Ren,
\newblock ``Underwater image enhancement via medium transmission-guided
  multi-color space embedding,''
\newblock {\em IEEE Transactions on Image Processing}, vol. 30, pp. 4985--5000,
  2021.

\end{thebibliography}

\end{document}